\documentclass[runningheads]{llncs}
\usepackage[utf8]{inputenc}

\usepackage{graphicx}
\usepackage{amssymb}
\usepackage{amsmath}
\usepackage{bbm}
\usepackage{pgfplots}

\usepackage{algorithm}
\usepackage[ruled,vlined,algo2e]{algorithm2e}

\usepackage{tikz}
\usetikzlibrary{shapes,arrows,positioning}
\usetikzlibrary{er,positioning,bayesnet}

\usepackage{caption} 
\usepackage{subcaption} 
\tikzset{
block/.style={
  draw, 
  fill=white, 
  rectangle, 
  minimum height=3em, 
  minimum width=6em,
  minimum width={width("Evaluate train")+15pt},
  },
sum/.style={
  draw, 
  fill=white, 
  circle, 
  cross,
  minimum size=1em
  },
input/.style={coordinate},
output/.style={coordinate},
}

\definecolor{blue}{HTML}{1F77B4}
\definecolor{orange}{HTML}{FF7F0E}
\definecolor{green}{HTML}{2CA02C}

\pgfplotsset{compat=1.14}

\begin{document}
\title{A Bayesian Neural Network\\ based on Dropout Regulation}

\author{Claire Theobald,  Frédéric Pennerath, Brieuc Conan-Guez, \\ Miguel Couceiro, Amedeo Napoli}

\authorrunning{C. Theobald et al.}

\institute{Université de Lorraine, CentraleSupélec, CNRS, Inria, LORIA, F-54000 France \email{}
}
\maketitle              
\section{Introduction}
\label{section:Introduction}
Deep learning is reaching very high levels of performance in many research domains, such as computer vision \cite{DBLP:conf/nips/KrizhevskySH12}, NLP \cite{DBLP:journals/corr/Graves13}, speech processing \cite{DBLP:journals/corr/abs-1303-5778}, and so on. In addition, deep learning is used in numerous application domains such as astrophysics, autonomous driving, biology, chemistry, medicine, where precision and uncertainty measurements are critical  \cite{crash}. At present, most of the neural networks are outputting point estimates of their predictions and do not always offer an accurate representation of their predictive uncertainty. For example, in computer vision, images are complex data which can induce a high amount of uncertainty. Then, pattern recognition is a critical task especially when it is related to decision making.

Over the past few years, research has been conducted on Bayesian Neural Networks (BNN), which consist of assuming a prior over the weights of the network instead of point estimates \cite{hinton_keeping_nodate}. This allows for the quantification of not only aleatoric uncertainty, which is the uncertainty related to the data, but of the epistemic uncertainty as well, the uncertainty of the model. In particular, MC Dropout is a BNN using dropout as a Bayesian approximation of the posterior distribution of the parameters of the network \cite{gal_dropout_2016}. In the MC dropout implementation, the dropout rate is fixed to a certain value. As such it is seen as a hyperparameter. However, there were several other attempts at trying to learn an optimal value of the dropout rate. This generally means computing the gradient with respect to the dropout rate which is a complex problem. Concrete Dropout \cite{gal_concrete_2017} solves this by using a relaxation of the Bernoulli Distribution, while Variational Dropout \cite{kingma_variational_2015} uses a Gaussian approximation, and LBD \cite{boluki_learnable_2020} uses an ARM estimator of the Bernoulli gradient. In any case, these methods are relatively complex to use. 

In this paper, we present a new method called ``Dropout Regulation'' (DR) inspired by automation and control loops. The goal is to regulate the dropout rate given the gap of performance on the training data and the validation data. The idea is that when the model has similar performances between the train and validation data, dropout is not needed, but when the model starts performing much better on the training data, we need a higher dropout rate for a stronger regularization. The dropout regulation try to automatically set up a dropout rate. This regulation method provides a precise estimation of the predictive uncertainty while being simple to implement. While it still requires the tuning of some controller coefficients, seen as hyperparameters, it does not require a gradient computation with respect to the dropout rate contrasting with other methods such as Concrete Dropout or LBD. The first part of this paper recalls principles of Bayesian deep learning and uncertainty estimation. The second presents the Dropout Regulation method. Finally, in the last part we will evaluate the ability of our method to estimate uncertainty and compare it with the state-of-the-art.

\section{Bayesian Deep Learning and uncertainty estimation}
\label{section:summary}

 \subsection{Epistemic and aleatoric uncertainties}
 \label{subsection:uncertainties}
   
   In a Bayesian machine learning model, the parameters\footnote{We denote model parameters as $w$ instead of adopting classical notation $\theta$ as these will shortly refer to weights of neural network.} $w \in \Omega$ are subject to a distribution $p(w)$, called the prior. The goal is then, given the dataset $\mathcal{D} = \{(x_i,y_i)\}_{i=1}^N \subset \mathcal{X} \times \mathcal{Y}$ - where $\mathcal{X}$ is the set of instances and $\mathcal{Y}$ the set of labels - to find the posterior distribution $p(w|\mathcal{D})$, using Bayes' rule:
   \begin{equation}
   p(w|\mathcal{D}) = \frac{p(\mathcal{D}|w)p(w)}{p(\mathcal{D})}
       \label{eq:bayes}
   \end{equation}
   $p(\mathcal{D}|w)$ is called the likelihood and $p(\mathcal{D})$ the model evidence. Given the posterior, and an instance $x$, one can then find the posterior predictive distribution, $p(Y|x,\mathcal{D})$ by marginalizing over the posterior:
   \begin{equation}
   p(Y|x,\mathcal{D}) = \int p(Y|x,w)p(w|\mathcal{D})dw
       \label{eq:marginal}
   \end{equation}
   This distribution tells us which outputs $y$ are the most likely for the instance $x$ given the data. However given the probabilistic nature of the model, one can also quantify the level of uncertainty of the model w.r.t. its predictions.
   
   The uncertainty of a model prediction can be decomposed into two types of uncertainties, namely epistemic and aleatoric uncertainty \cite{kendall_what_nodate}.
   
   \begin{itemize}
         \item Aleatoric uncertainty is the uncertainty that is inherent to the data. Whether it be random noise or ambiguous data, this uncertainty can be explained by the fact that generally, the perfect oracle $f(x)$ which predicts correctly every time all possible instances $x$ is non-deterministic.
       \item On the other hand, epistemic uncertainty, or model uncertainty, represents the lack of knowledge the model has on the data. This is often due to a lack of data, either in amount or diversity, or a training that was not long enough for the model to learn sufficient information about the data. Generally this uncertainty is reducible.
   \end{itemize}
   
   The total predictive uncertainty of the model is the sum of those two uncertainties. In the case of classification, the epistemic and aleatoric uncertainties can be evaluated using information theory as \cite{DBLP:conf/icml/DepewegHDU18}:
  \begin{align}
   &\mathcal{U}_{aleatoric} (x) = \mathbb{H}(Y|x,W, \mathcal{D}) = \mathbb{E}_{p(w|\mathcal{D})}[\mathbb{H}(Y|x,w)] \label{eq:dataunc}\\
   &\mathcal{U}_{model} (x) = \mathbb{I}(Y,W |x, \mathcal{D}) =  \mathbb{E}_{p(w|\mathcal{D})}\left[KL(p(Y|x,w) || p(Y|x, \mathcal{D}))\right] \label{eq:modunc}
   \end{align}
   
   The total predictive uncertainty being the sum of the two uncertainties:
   \begin{equation}
       \mathcal{U}_{pred} (x) = \mathbb{I}(Y,W |x, \mathcal{D}) + \mathbb{H}(Y|x,W, \mathcal{D}) = \mathbb{H}(Y|x, \mathcal{D})
       \label{eq:predunc}
   \end{equation}
   Equation \ref{eq:dataunc} means that the aleatoric uncertainty is the mean predictive entropy over the posterior. Quantitatively, this uncertainty will be high if the entropy is high on average, which means that each sampled predictive posterior are, on average, uncertain.
    
    Equation \ref{eq:modunc} is literally the mutual information between the prediction $Y$ and the parameters $W$. It can also be interpreted as the mean KL divergence between each sampled posterior predictive with the average posterior predictive. Therefore this uncertainty will be high if the distribution $p(Y|W, x)$ varies a lot for different parameters $w$.

\subsection{Using dropout as a Bayesian approximation in Bayesian Neural Networks}

Given a training dataset $\mathcal{D} = \{(x_i, y_i)\}_{i=1}^N \subset \mathcal{X} \times \mathcal{Y}$, a neural network is a function $f_w : \mathcal{X} \rightarrow \mathcal{Y}$ parametrized by $w \in \Omega$ which associates to each instance $x$ a prediction $\hat{y}=f_w (x)$.

More precisely, a neural network is defined with $L$ hidden layers, $l \in \{1, ..., L\}$ the index of the hidden layers, with $y^l$ the output of layer $l$ (with $y^0=x$ and $y^L=\hat{y}=f_w(x)$), $w^l$ and $b^l$ the weight and bias matrices at layer l. For input $x$, the output $\hat{y}$ can be calculated by the following recurring equations, for $l \in \{1, ..., L-1\}$:
 \begin{equation}
y^{l+1}=\sigma\left(w^{l+1} y^l + b^{l+1}\right)
\label{eq:nn}
\end{equation}
With $\sigma$ being an activation function, usually a non linear function such as the sigmoid $\sigma(x)=\frac{1}{1+\exp(-x)}$.

Here, $w=\{(w^l,b^l)\}_{l=1}^L$ are the parameters of the neural network which can be trained to optimize the predictive function $f_w$.

Dropout has been first proposed as a regularization technique \cite{srivastava_dropout_nodate}, i.e. a way to slow down and prevent the overfitting of the data in neural networks. Overfitting can be described as a difference in performance between the training dataset $\mathcal{D}^{train}$ and the test dataset $\mathcal{D}^{test}$. When the neural network starts performing very well on $\mathcal{D}^{train}$ at the cost of reduced accuracy on $\mathcal{D}^{test}$, overfitting is happening: the neural network is losing the ability to generalize to a wider set of unseen data. When using Dropout on a neural network, the outputs of each layer $y^l$ are multiplied by a binary noise. The recurrence operation becomes then:
 \begin{align}
&r^l_j \sim Bernoulli(1-p^l) \label{eq:dropout} \\
&y^{l+1} = \sigma\left(w^{l+1} (r^{l+1} * y^l) + b^{l+1}\right) \label{eq:dropout_nn}
\end{align}
Where $*$ is the Hadamard (element-wise) product.

The dropout rates $p=\{p^l\}_{l=1}^L$ are hyperparameters that are not part of $w$. In most applications, the dropout rates are set arbitrarily to a certain value (often 0.5), or through a grid search.

In Bayesian Neural Networks, the parameters of the network $w$ are drawn from a prior distribution $p(w)$ instead of a point estimate, which makes it a Bayesian model as seen in Section \ref{subsection:uncertainties}. As the true posterior $p(w|\mathcal{D})$ is very often intractable, the goal instead is to optimize an approximate posterior $q_{\theta}(w)$ parametrized by $\theta$ by minimizing the KL-divergence between $q_\theta(w)$ and $p(w|\mathcal{D})$:
   \begin{equation}
   KL(q_\theta(w) || p(w|\mathcal{D}))
   \label{eq:KL}
   \end{equation}
   In MC Dropout \cite{gal_dropout_2016}, the approximate posterior is defined as a Bernoulli distribution over the weights, using Dropout. In Concrete Dropout \cite{gal_concrete_2017}, the dropout rate is optimized by using a relaxed distribution which approximates the Bernoulli distribution into a continuous space. The training of a BNN is then almost identical to a deterministic network, the difference being that dropout is being used at testing time to assess the uncertainty of the predictions. 
   
   In order to evaluate the posterior predictive distribution $p(y|x,\mathcal{D})$, we need to integrate over the posterior $p(w|\mathcal{D})$. That is usually not possible. Instead, we apply Monte Carlo (MC) sampling: we draw $K$ samples from the posterior: $(w_k)_{k=1}^K \sim q_{\theta}(w)$, and retrieve each sampled approximate posterior predictive distribution:
\begin{equation}
   p(y|x,\mathcal{D})=\int p(y|x,w)p(w|\mathcal{D})dw \approx \int p(y|x,w)q_\theta(w)dw \approx \frac{1}{K} \sum_{k=1}^K p(y|x,w_k)
   \label{eq:pred}
   \end{equation}
So, by using dropout as a Bayesian approximation for neural networks, we are able to turn them into a Bayesian model, which allows us to compute not only aleatoric uncertainty but also epistemic uncertainty, as seen in Section \ref{subsection:uncertainties}, in Equations \ref{eq:dataunc}, \ref{eq:modunc} and \ref{eq:predunc}. Now, MC Dropout raises a new constraint: $p$ being a hyperparameter, how to select a good value for it while having a precise estimation of the uncertainties ? One can select it arbitrarily, or through a grid search. Several papers proposed to include $p$ as a variational parameter, whether by approximating it by a continuous distribution \cite{gal_concrete_2017}  \cite{kingma_variational_2015} or using an estimator of the gradient \cite{boluki_learnable_2020}. We are going to present in Section \ref{section:dropout_reg} a new method to find the dropout rate, Dropout Regulation. This method is not based on a variational approach for $p$ and more simple to implement compared to state-of-the-art, while still precisely estimating the predictive uncertainty.
    
\section{Dropout regulation}
\label{section:dropout_reg}
\subsection{General presentation}

In order to find the best parameters $w^*$, the neural network has to solve an optimization problem. First, we need to define a loss function $L: \mathcal{Y} \times \mathcal{Y} \rightarrow \mathbb{R}$ that measures how good the model predicts the label $y$ of an instance $x$. The goal of the model is then to minimize the risk $\mathcal{R} : \Omega \rightarrow \mathbb{R}$ or expected loss:
 \begin{equation}
\mathcal{R}(w) = \int_{\mathcal{X} \times \mathcal{Y}} L(f_w(x),y) dP(x,y)
\label{eq:risk}
\end{equation}
$P(x,y)$, the underlying distribution generating the data, is generally unknown. Therefore, a common substitute to the risk is the empirical risk:
 \begin{equation}
\hat{\mathcal{R}}^{train} (w) = \frac{1}{N} \sum_{(x,y) \in \mathcal{D}^{train}} L(f_w(x),y)
\label{eq:emp_risk}
\end{equation}
The goal of the training process is therefore to find the optimal solution to the empirical risk given the training data:
 \begin{equation}
w^* = \arg\min_{\Omega} \hat{\mathcal{R}}^{train}(w)
\label{eq:opt_theta}
\end{equation}
Now, the empirical risk is defined on $\mathcal{D}^{train}$, but an empirical risk $\hat{\mathcal{R}}^{val}$ can also be defined on validation data, i.e. a set $\mathcal{D}^{val} \subset \mathcal{X} \times \mathcal{Y}$ such that $\mathcal{D}^{val} \cap \mathcal{D}^{train} = \emptyset$ or at least $\mathcal{D}^{val} \neq \mathcal{D}^{train}$. Such risk will not be used to find $w^*$, but can be used to check if $w^*$ is an appropriate solution that can enable the model to generalize to a broader set of data.

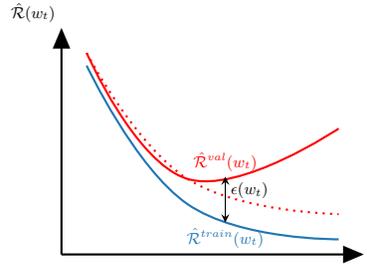
\begin{figure}
    \centering
    \begin{tikzpicture}[scale=0.67, every node/.style={scale=0.67}]
      \draw[thick,->] (0,0) -- (6,0) node[anchor=north west] {$t$};
      \draw[thick,->] (0,0) -- (0,4.5) node[anchor=south east] {$\hat{\mathcal{R}}(w_t)$};
      \draw[red, thick] plot [smooth, tension=0.75] coordinates {(0.5,4) (2.5,1.5) (5.5,2.5)};
      \draw[red, dotted, thick] plot [smooth, tension=0.75] coordinates {(0.5,4) (2.5,1.5) (5.5,0.8)};
      \draw[blue, thick] plot [smooth, tension=0.75] coordinates {(0.5,3.75) (2.5,1) (5.5,0.3)};
      \node[blue, name=train] at (3.25,0.33) {$\hat{\mathcal{R}}^{train}(w_t)$};
      \node[red, name=val] at (3.25,1.85) {$\hat{\mathcal{R}}^{val}(w_t)$};
      \draw[<->,>=stealth] (train) -- node[pos=0.7, right] {$\epsilon(w_t)$} (val);
    \end{tikzpicture}
    \caption{Illustration of overfitting. As the training process progresses, the gap $\epsilon(w_t)$ between the validation set and the training set increases, reducing the generalization capability of the model. The red line represents the empirical risk on the validation data, the blue line the empirical risk on the training data. The dotted line is a desired behaviour of the empirical risk on the validation data without overfitting.}
    \label{fig:overfitting}
\end{figure}

Generally, the validation empirical risk is greater than the training empirical risk, i.e. $\hat{\mathcal{R}}^{train}(w) \leq \hat{\mathcal{R}}^{val}(w)$. An ideal scenario would be to have an equality, meaning that the model performs just as good on the training dataset than on the validation dataset. However, what happens oftentimes is that $\hat{\mathcal{R}}^{train}(w)$ decreases to an optimum during training and $\hat{\mathcal{R}}^{val}(w)$ increases as the model loses the ability to generalize. Therefore we want the quantity $\epsilon(w)=\hat{\mathcal{R}}^{val}(w)-\hat{\mathcal{R}}^{train}(w)$ to stay as close to zero as possible, or prevent it to increase too much. This gap $\epsilon(w)$ between the two empirical risks can be considered as a measure of overfitting (Figure \ref{fig:overfitting}).

Regularization techniques have been developed to try to reduce the gap as much as possible to keep the best performance possible on the validation or test dataset. Such techniques include weight decay and dropout. We will focus on dropout in this article. As seen in Section \ref{section:summary}, there is a lot of discussions around the ``best'' value of the dropout rate. Indeed, while a fixed dropout rate is sufficient, it could prevent the model to learn more efficiently if the gap $\epsilon(w)$ is low enough, and it may not be enough to prevent overfitting if $\epsilon(w)$ is already increasing.

The work in this article is therefore inspired by automation and controlled loops. Indeed, we want to adjust the dropout rate dynamically during training with respect to the gap $\epsilon(w)$. Here we will assume a single value $p$ for the dropout rate for all layers. Algorithm \ref{alg:alg_1} describes how to adjust the dropout rate during training.

\begin{figure}
\centering
\begin{tikzpicture}[auto,>=latex',cross/.style={path picture={ 
  \draw[black]
(path picture bounding box.south east) -- (path picture bounding box.north west) (path picture bounding box.south west) -- (path picture bounding box.north east);
}}]
    \node [input, name=input] {};
    \node [sum, cross, right = of input] (sum) {};
    \node [block, right = of sum] (controller) {Dropout controller};
    \node [block, right = of controller] (model) {Learn/Fit};

    \node [output, right = 11cm] (output) {};
    \node [output] at (9cm,-1.75cm) (eval1) {};
    \node [output] at (9cm,-3.75cm) (eval2) {};
    \node [block] at (7.7cm,-1.75cm) (eval_t) {Evaluate train};
    \node [block] at (7.7cm,-3.75cm) (eval_v) {Evaluate val};
    \node [sum, cross] at (4cm,-2.75cm) (sum2) {};
    
    \draw [->] (controller) -- node[name=v] {$p$} (model);
    \draw [->] (model) -- node[name=u] {$w$} (output);
    \draw [->] (u) -- ++(0,-1.94cm) -- (eval1);
    \draw [->] (9.89cm,-1.75cm) -- ++(0,-2cm) -- (eval2);
    \draw [->] (eval_t) -| node[pos=0.2, above] {$\hat{\mathcal{R}}^{train}(w)$} node[pos=0.95, left] {$-$} (sum2);
    \draw [->] (eval_v) -| node[pos=0.2, above] {$\hat{\mathcal{R}}^{val}(w)$} node[pos=0.95] {$+$} (sum2);
    \draw [->] (sum2) -| node[pos=0.335] {$\hat{\mathcal{R}}^{val}(w)-\hat{\mathcal{R}}^{train}(w)$} node[pos=0.98] {$+$} (sum);
    \draw [draw,->] (input) -- node[pos=0.90] {$-$} node[pos=0] {$\epsilon_0=0$} (sum);
    \draw [->] (sum) -- node {$\epsilon(w)$} (controller);

\end{tikzpicture}
\captionof{figure}{Control loop representation of Algorithm \ref{alg:alg_1}}
\label{fig:fig_control_1}
\end{figure}
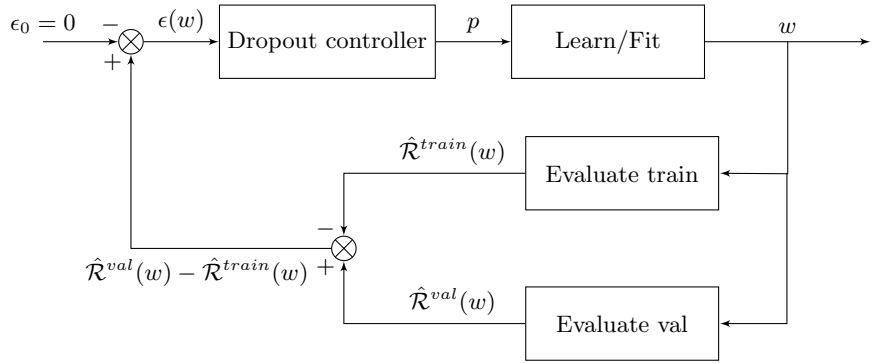

In Figure \ref{fig:fig_control_1}  it can be seen how the dropout regulation algorithm is reminiscent of a control loop. The target value $\epsilon_0=0$ is on the left. The quantity that we want to control and possibly set to 0 is $\delta\epsilon=\epsilon(w)-\epsilon_0=\epsilon(w)$ which is represented by the sum-operator on the left of the figure, the symbol of a cross inside a circle with the signs of the inputs outside. $\epsilon(w)$ then goes through the dropout controller to compute the next dropout rate $p=p(\epsilon(w))$ (Dropout Controller block). After that, the model learns through a new epoch with the new dropout rate to compute the updated parameters $w$ (Learn/Fit Block). At this point, the algorithm can stop at any fixed point to output $w$ (arrow on the far right). $w$ is also then used to evaluate the empirical risks of both the training and validation datasets (Evaluate train/val blocks below). These risks are then subtracted to get $\epsilon(w)$ (sum-operator below). We then loop back by computing $\delta\epsilon$. Note that in Algorithm \ref{alg:alg_1} we select $w$ such that the empirical risk on the validation data is minimal.


\begin{algorithm}
\SetAlgoLined
~\\
\textbf{Input:} Initial parameters $w_0$, training dataset $\mathcal{D}^{train}$, validation dataset $\mathcal{D}^{val}$, dropout controller function $p(\epsilon): \mathbb{R} \rightarrow [0,1]$, number of epochs $T$ \\
\textbf{Output:} optimal parameters $w_{t^*}, p_{t^*}$\\
Set $p_0=0$ \\
  \For{$t=1$ to $T$}{
    $w_{t}=Fit(w_{t-1})$  w.r.t. $p_{t-1},\mathcal{D}^{train}$ \\
     Compute $\epsilon(w_t) =\hat{\mathcal{R}}^{val}(w_t)-\hat{\mathcal{R}}^{train}(w_t)$\\
    Set $p_{t}=p(\epsilon(w_t))$
 }
  Return optimal parameters $w_{t^*}, p_{t^*}$ s.t. $t^* = \arg\min_{t} \hat{\mathcal{R}}^{val}(w_t)$
\captionof{figure}{Dropout regulation training}
 \label{alg:alg_1}
\end{algorithm}

\subsection{PI dropout controller}

 The representation of the algorithm as a control loop allows us to consider the system as a form of controller, with the hope of adjusting $p$ to set $\epsilon(w)$ to the desired value of 0. All that needs to be done now is how to specifv $p(\epsilon)$. We will use for this application the PID controller.
 
 The PID (Proportional-Integral-Derivative) controller is defined as follows. Let $y(t)$ be a measured process variable, $r(t)$ a desired setpoint for $y(t)$, $\epsilon(t)=y(t)-r(t)$ the error measured between the process variable and the target. Let $u(t)$ be a control variable that can be adjusted by the controller over time to minimize the error. A PID controller sets $u(t)$ to be:
  \begin{equation}
  u(t) = K_p \left( \epsilon(t) + \frac{1}{T_i} \int_0^t \epsilon(\tau)d\tau + T_d \frac{d\epsilon(t)}{dt}\right)
   \label{eq:PID_2}
   \end{equation}
With $K_p, T_d \geq 0, T_i >0$.

In Dropout Regulation, we compute $u(t)$ using Equation \ref{eq:PID_2} by using the error $\epsilon(t)$ between the two risks. However $u(t) \in \mathbb{R}$ and the control variable that the network can use is the dropout rate $p$ which is in $[0,1]$. Remind that dropout is a form of regulation. When the error is low, or even negative, it means that the network needs to learn from its training data better and therefore dropout is not necessary. If the error is too high it means that the model is overfitting and dropout is needed. As $\epsilon \rightarrow +\infty$, $p(\epsilon) \rightarrow 1$. Therefore $p(\epsilon(w))$ is defined as follows:
 \begin{equation}
     p(\epsilon(w)) = (1-\exp(-u(t)))\mathbbm{1}_{\mathbb{R}^+}(u(t))
     \label{eq:P_cont}
 \end{equation}
 Where $\mathbbm{1}_{\mathbb{R}^+}(x)=1$ if $x\geq0$, $0$ otherwise.
 We have now defined every parameter of our controller. We will only consider a PI controller, i.e. $T_d=0$. Also considering that time is discretized, the integral action is approximated by a sum. Ultimately, our controller is defined as:
   \begin{equation}
  u(t) = K_p \left( \epsilon(w_t) + \frac{1}{T_i} \sum_{k=1}^t \epsilon(w_k)\right)
   \label{eq:PID_3}
   \end{equation}
And $p(\epsilon(w_t))$ as defined in Equation \ref{eq:P_cont}. The entire regulation process is described in Algorithm \ref{alg:alg_2}. 

\begin{algorithm}
\SetAlgoLined
~\\
\textbf{Input:} Parameters $w$, training dataset $\mathcal{D}^{train}$, validation dataset $\mathcal{D}^{val}$, $K_p$, $T_i$, number of epochs $T$ \\
\textbf{Output:} optimal parameters $w_{t^*}, p_{t^*}$\\
Set $p_0=0$ \\
  \For{$t=1$ to $T$}{
    $w_{t}=Fit(w_{t-1})$  w.r.t. $p_{t-1},\mathcal{D}^{train}$ \\
     Compute $\epsilon(w_t) =\hat{\mathcal{R}}^{val}(w_t)-\hat{\mathcal{R}}^{train}(w_t)$\\
    Compute $u(t)= K_p \left( \epsilon(w_t) + \frac{1}{T_i} \sum_{k=1}^t \epsilon(w_k)\right)$\\
    Set $p_{t+1}=(1 - \exp(-u(t))) \mathbbm{1}_{\mathbb{R}^+}(u(t))$ 
 }
 Return optimal parameters $w_{t^*}, p_{t^*}$ s.t. $t^* = \arg\min_{t} \hat{\mathcal{R}}^{val}(w_t)$
\captionof{figure}{PI dropout controller training}
 \label{alg:alg_2}
\end{algorithm}

   \section{Experiments}
   \subsection{Setup}
   
   The PI dropout controller has been implemented in the context of an image classification problem. The dataset used was the Dogs vs. Cats Kaggle dataset \cite{dogs_vs_cats}. We are using 8000 images for training, 1600 for validation and 8000 for testing. The architecture is based on VGG-16 \cite{simonyan_very_2015} with all weights preloaded and freezed but the last convolution block. Then, the output of the convolution part is flattened, and two dense layers of 256 neurons with dropout are added before the last output layer.
   
   The loss used is the categorical crossentropy, using a SGD optimizer with learning rate $10^{-4}$, momentum 0.9 and a batch size of 100, and 60 epochs. We are comparing 4 methods: Regulation Dropout with $K_p=10$ and $T_i=10000$, Concrete Dropout, MC Dropout and a deterministic network. For the deterministic network, dropout is not used at testing time and therefore only the aleatoric uncertainty is used for uncertainty measurements. At testing time, we compare the models which had the best validation loss during training.
   
    In order to assess the quality of the uncertainty measured, we are using the PAvPU metric, which combines the probability that a prediction that is certain is accurate: $P(accurate | certain)$, and the probability that an inaccurate prediction is uncertain: $P(uncertain | inaccurate)$ \cite{mukhoti_evaluating_2019}. The PAvPU metric is therefore calculated as such:
     \begin{equation}
   PAvPU = \frac{n_{iu}+n_{ac}}{n_{iu}+n_{ac}+n_{ic}+n_{au}}
   \label{eq:pred}
   \end{equation}
   
   Where $n_{iu}$ is the number of inaccurate and uncertain samples, $n_{ac}$ the number of accurate and certain samples, $n_{ic}$ the number of inaccurate and certain samples and $n_{au}$ the number of accurate and uncertain samples.
   
   The PAvPU metric is evaluated at different thresholds of uncertainty, ranging from $u_{min}$ to $u_{max}$ in a linear progression, where $u_{min}$ and $u_{max}$ are respectively the minimum and maximum uncertainty over the test dataset:
     \begin{equation}
   u_{thres}(t) = u_{min} + t(u_{max}-u_{min}), t \in [0,1]
   \label{eq:pred}
   \end{equation}

   \subsection{Results}
   
   Table \ref{tab:results} shows the accuracy and PAvPU metric for the different methods tested. The PAvPU metric was evaluated at an uncertainty threshold equal to the mean uncertainty over the test dataset. The accuracy of Regulation is slightly below of Concrete but above Deterministic and MC Dropout. 
   
   We can see that for all methods, the accuracy is very high and so the number of accurate samples is very high when compared to the inaccurate. That means that a higher uncertainty will decrease the number of accurate and certain samples. So, interestingly, the PAvPU value for the deterministic network is greater than that of MC Dropout. This is because MC Dropout represents more uncertainty overall (because of the added epistemic uncertainty). Then some accurate samples are classified as uncertain although they were not in the deterministic method. Because the proportion of inaccurate samples is so low, the added uncertainty is not compensated by the number of inaccurate and uncertain samples. Both Concrete and Regulation show similar levels of PAvPU, so our method is quite competitive.

   Figure \ref{fig:pavpu} shows the results of the PAvPU metric curve for varying thresholds of uncertainty, ranging from $t=0$ to $t=1$. We can observe that Regulation is quite similar to Concrete in terms of PAvPU metric performance, and that they are both above Deterministic and MC Dropout. The PAvPU curve for MC Dropout is below the curve of Deterministic, which relates to the previous explanation.

  \begin{minipage}{\textwidth}
  \begin{minipage}[b]{0.49\textwidth}
  \centering
\begin{tabular}{|l|l|l|}
\hline
Method &  Accuracy & PAvPU \\
\hline
Deterministic &  0.932 & 0.7155 \\
MC Dropout & 0.928 & 0.70734\\
Concrete & {\bfseries 0.935} & 0.7575\\
Regulation & 0.934 & {\bfseries 0.7579}\\
\hline
\end{tabular}
\captionof{table}{Accuracy and PAvPU metrics for the different methods. Bold indicates the best metric obtained by the method used.}\label{tab:results}
    \end{minipage}
     \begin{minipage}[b]{0.49\textwidth}
        \centering
        \includegraphics[scale=0.34]{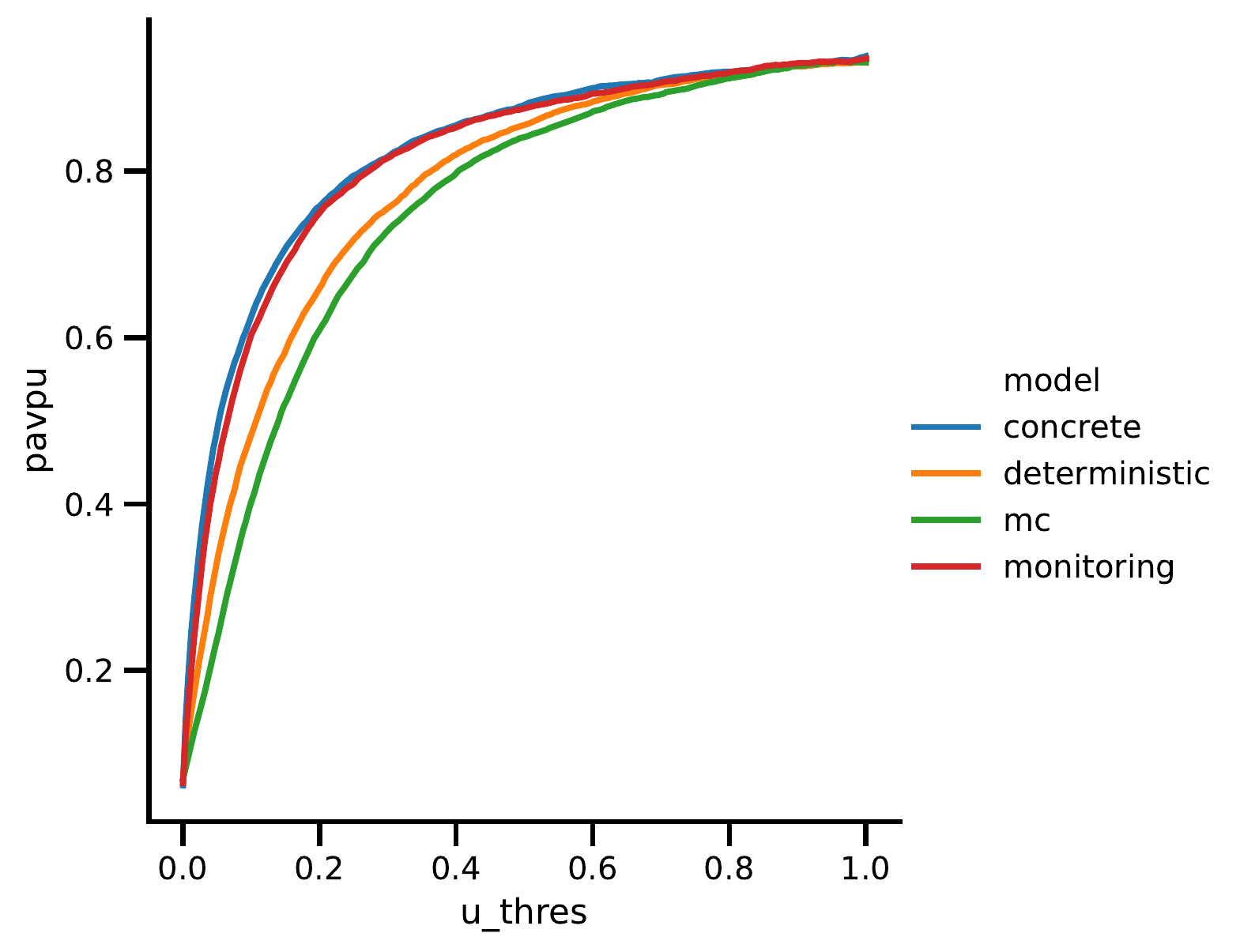}
        \captionof{figure}{PAvPU metric for 250 different thresholds on uncertainty and different models.}
        \label{fig:pavpu}
     \end{minipage}
  \end{minipage}

  \subsection{Influence of $K_p$}
 
We will now study the influence of the hyperparameter $K_p$ of the controller. This parameter influences the strength of the adjustment of the dropout rate $p$. The higher $K_p$ is, the higher $p$ will become for a given error $\epsilon(w)$. We are going to compare two different values of $K_p$. First of all, we are keeping the same parameters as the previous experiment, but this time we have $N_{train}=N_{test}=1000$ training and test images and $N_{val}=200$ validation images and a SGD optimizer with learning rate $10^{-4}$ with no momentum. The two values of $K_p$ that we will compare are $K_p=0.15$ and $K_p=0.50$. These values of $K_p$ are lower when compared to the previous experiment because of the absence of momentum. Indeed this parametrization allows us to spotlight an interesting oscillating behavior of the controller. We compare two models: one model with only 20\% of the data and another with 100\% of the data.

\begin{figure}
\centering
\begin{subfigure}{.5\textwidth}
  \centering
  \includegraphics[width=.80\linewidth]{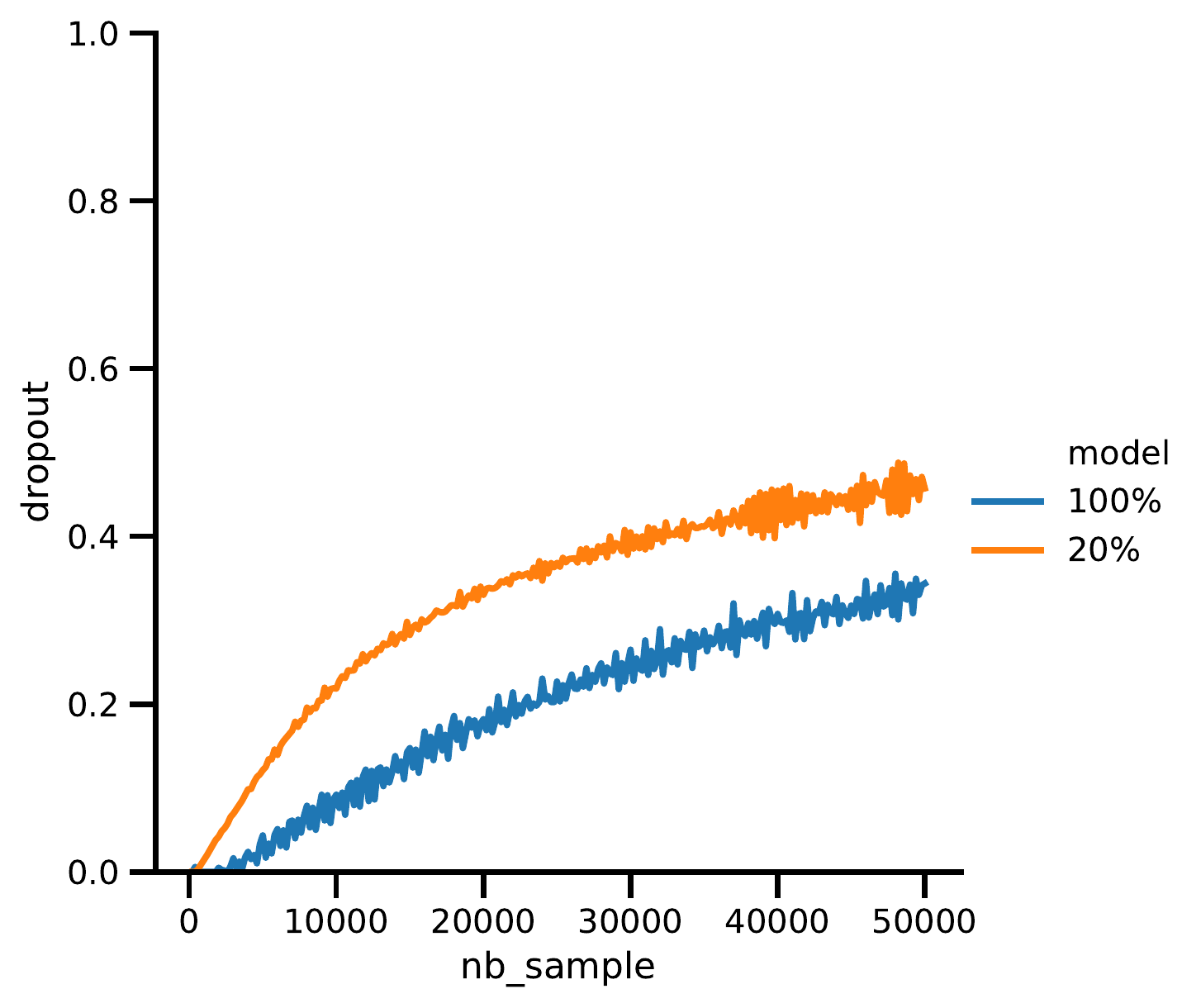}
  \caption{}
  \label{fig:kp1}
\end{subfigure}%
\begin{subfigure}{.5\textwidth}
  \centering
  \includegraphics[width=.80\linewidth]{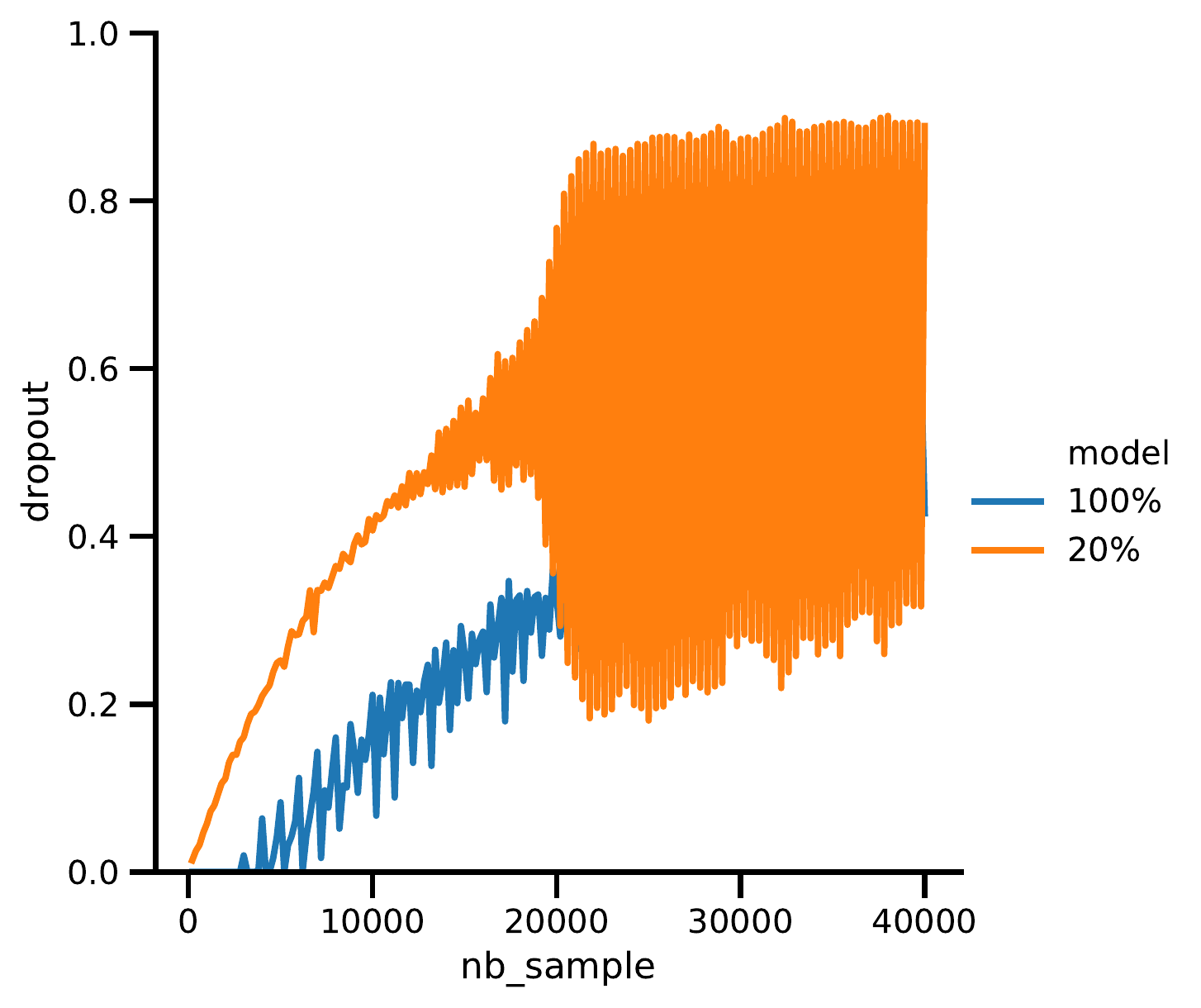}
  \caption{}
  \label{fig:kp2}
\end{subfigure}
\caption{Dropout rate for $K_p=0.15$ (a) and $K_p=0.50$ (b). The x-axis represents the amount of images seen during training. It is equivalent to the number of epochs multiplied by the number of images per epoch. In blue the dropout rate for the model with 20\% of the data, in orange 100\% of the data.}
\end{figure}

As one can see, an interesting behavior happens when we increase $K_p$. In both cases, the dropout starts by slowly increasing, reflecting the increase of $\epsilon(\omega)$ as both performances on the validation and trains datasets diverge. We can also see that the dropout rate is found to be lower for the largest dataset, which is reasonable because the amount of data is in itself a form of regularization. In Figure \ref{fig:kp1}, the dropout rate is more or less stable throughout. However, in Figure \ref{fig:kp2}, we observe a divergence of the dropout rate as soon as the value of the dropout rate reaches approximately 0.5, for the model with $20\%$ of the data. Indeed, when the dropout rate starts to reach higher values, the network has less and less active neurons to learn complex patterns which may be necessary to learn from the data well. Therefore, the performance on the train and validation datasets both start to plummet, to the point where the model basically starts to guess randomly. Because of that, the error $\epsilon(\omega)$ indirectly reduces since the performances on both datasets are almost equally worse. Which means that the controller will select next a dropout rate which is way lower, in general $p<0.5$. Since most of the weights were ``frozen'' when the dropout rate was high, when the dropout rate becomes lower again, the model performs well again and $\epsilon(\omega)$ is high again, which means that on the next epoch the dropout rate will be very high, etc. It is actually very interesting to see that such a behavior happens when we apply a controller on a neural network. In most physical systems where a controller can be used, such oscillations can be encountered when the parameters of the controller are too strong. Here also, one must choose the hyperparameters of the controller well to prevent such behavior from happening. Other hyperparameters like the optimizer (momentum) or the size of dataset can also be influential. In any case, since we select in the end the model which has the best performance on the validation data overall, the influence of the oscillations - if they happen - on the final performance are mitigated.

   \section{Conclusion}
   
   In this paper, we presented a method for adjusting the dropout rate during training, called Dropout Regulation. By using control loops from automation, the goal was to empirically adjust the dropout rate in order to have a precise estimation of the uncertainty while being simple to implement. We then applied this method to an image classification problem. Our method got comparable results when compared to other methods w.r.t. accuracy and uncertainty metrics. Even if our method requires the tuning of the hyperparameters of the controller which can induce some side effects like an oscillating system, Dropout Regulation offers an alternative to the state of the art which makes it interesting. 
   
      \section{Acknowledgements}
      
      The first author is preparing a PhD thesis at the LORIA Lab in the context of the AstroDeep Research Project (\url{https://astrodeep.pages.in2p3.fr/website/projects/}) funded by ANR under the grant ANR-19-CE23-0024.

%
%
%

\bibliographystyle{splncs04}



%
\end{document}